\documentclass{egpubl}
\usepackage{gch2023}
\usepackage{subcaption}
 \WsShortPaper      

\usepackage[T1]{fontenc}
\usepackage{dfadobe}  
\usepackage{mathtools}
\usepackage{cite}  
\BibtexOrBiblatex
\electronicVersion
\PrintedOrElectronic
\ifpdf \usepackage[pdftex]{graphicx} \pdfcompresslevel=9
\else \usepackage[dvips]{graphicx} \fi

\usepackage{egweblnk} 

\title[One-to-many Reconstruction of 3D Geometry of cultural Artifacts]{One-to-many Reconstruction of 3D Geometry of cultural Artifacts using a synthetically trained Generative Model}

\author[Thomas Pöllabauer, Julius Kühn, Jiayi Li, Arjan Kuijper]
{\parbox{\textwidth}{\centering T. Pöllabauer$^{1,2}$\orcid{0000-0003-0075-1181},
         J. Kühn$^{1}$\orcid{0000-0003-2458-0030}, 
         J. Li$^{2}$\orcid{0009-0009-6866-1722},
         A. Kuijper$^{1,2}$\orcid{0000-0002-6413-0061}
        }
        \\
{\parbox{\textwidth}{\centering $^1$Fraunhofer Institute for Computer Graphics Research IGD \\
                                $^2$Interactive Graphics Research Group, TU Darmstadt
      }
}
}

\begin{document}
\captionsetup{labelfont=bf,textfont=it}
\teaser{
}

\maketitle
\begin{abstract}
Estimating the 3D shape of an object using a single image is a difficult problem. Modern approaches achieve good results for general objects, based on real photographs, but worse results on less expressive representations such as historic sketches. Our automated approach generates a variety of detailed 3D representation from a single sketch, depicting a medieval statue, and can be guided by multi-modal inputs, such as text prompts. It relies solely on synthetic data for training, making it adoptable even in cases of only small numbers of training examples. Our solution allows domain experts such as a curators to interactively reconstruct potential appearances of lost artifacts.


\end{abstract}  




\section{Introduction}
Historically cultural artifacts have been lost due to factors like mishandling, fires, conflicts, natural disasters, deterioration, theft, etc. Reconstructing these artifacts is important to preserve our collective history, understand past cultures' accomplishments, and shed light on their artistic and technological advancements.\\
To assist in this endeavour, we propose a fully-automatic reconstruction of 3D geometry of cultural heritage artifacts based on as little as a single, low-quality sketch. We illustrate the concept on poorly preserved medieval sketches of red pigments, so called sinopia, painted on a wall and originally used by stucco workers to sketch out their work. 
We use pretrained state-of-the-art machine learning models to solve problems related to cultural heritage by fine-tuning with synthetically generated data, allowing the adoption of these algorithms within a new domain.
Our key contributions are:
\begin{itemize}
    \item Screening of the relevant literature for one-to-many generative reconstruction of 3D geometry with a focus on applicability to the cultural heritage domain.
    \item Illustrating the applicability of the state-of-the-art generative models such as GANs and diffusion models to the reconstruction of cultural heritage artifacts, using a variety of different inputs. We illustrate the process with simple photographs of sketches on a medieval wall.
    \item Proposal of an interactive authoring tool for domain experts to reconstruct potential appearances based on simple inputs such as images and text.
\end{itemize}

\section{Related Work}
\begin{figure}
	\includegraphics[width=0.49\textwidth]{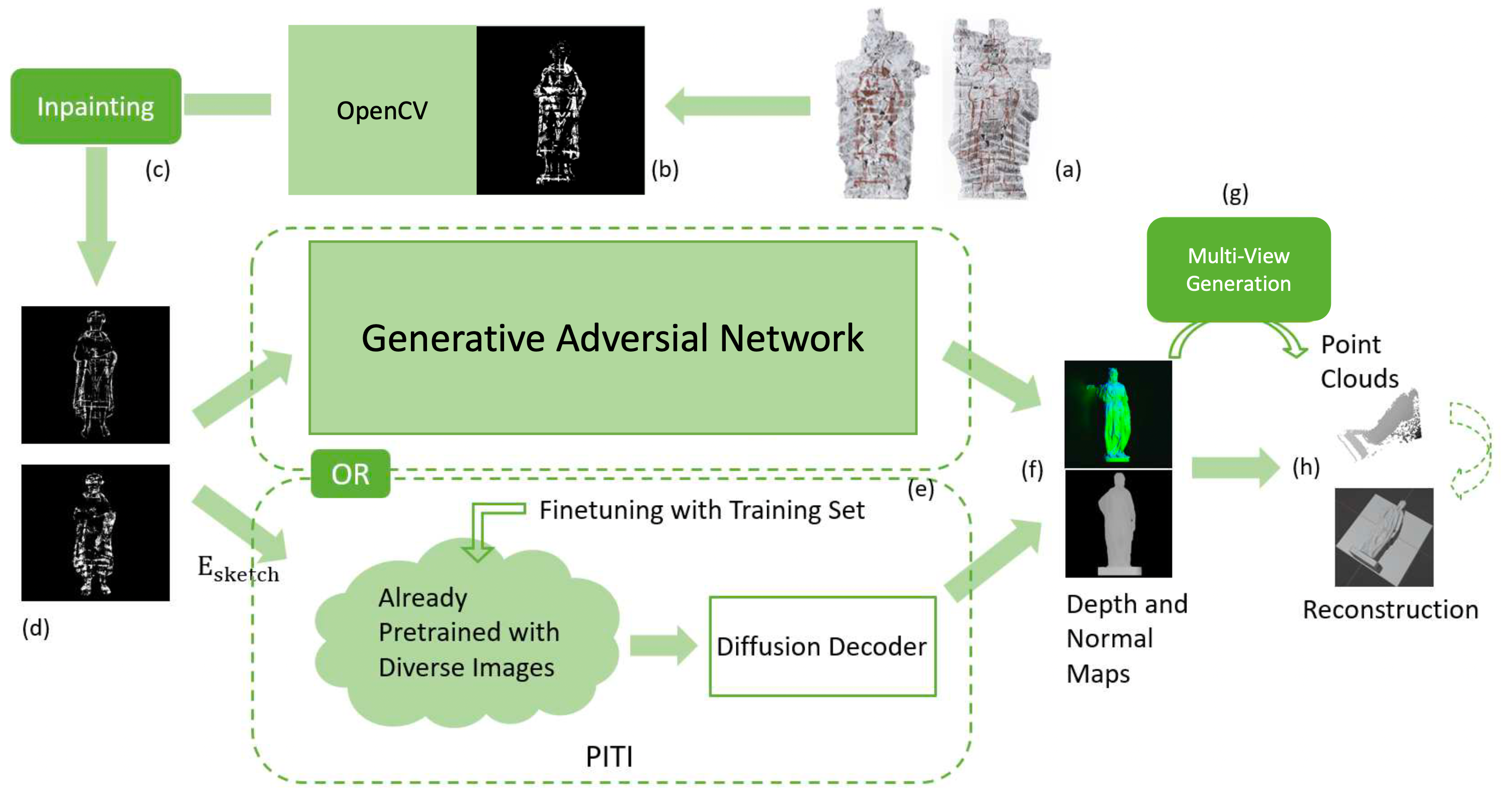}
	\caption
	{The proposed approach consists of the following components:
        (a) Input data, the original sinopia images.
        (b) Input image pre-processing with OpenCV.
        (c) Image inpainting  to restore the lines and repair holes.
        (d) Inpainted images serve as input for the diffusion model.
        (e) We use the PITI diffusion approach for further generation.
        (f) Depth map and normals are optained by the diffusion outputs.
        (g) An optional generation network is proposed to create more views to reconstruct multi-perspective 3D models. 
        (h) Point clouds and 3D shapes are generated from the depth and normal maps, and can be used for full geometry reconstruction.}
        \label{fig:approach}
\end{figure}
The reconstruction of 3D objects from images, ranging from highly expressive photographs, to drawings, and even simple line drawings, has been a topic of research in computer vision and computer graphics for a long time already. \\
Early non-machine learning-based methods for 3D geometry estimation from line drawings focused on local properties such as convexity, continuity, and faces. These methods can identify polyhedra and geometric shapes, but are less suitable for organic or natural objects or complex works of art. More recent works have proposed ways to create 3D surfaces from line drawings by exploiting specific geometric constraints, but require accurate input sketches. An example of this is True2Form, which produces relatively accurate 3D shapes, but requires precise input lines.
A classic approach to 3D reconstruction from images is photogrammetry, which achieves high quality results different scenes and objects, ranging from small statues to large areas, but requires multiple overlapping photographs and the scene to be static. \\
Recent 3D reconstruction approaches use deep learning methods. These approaches require a large amount of training data, but by adopting pre-training and synthetic data generation, state-of-the-art algorithms can be used even with limited available data, the major reason for low adoption in the domain of cultural heritage \cite{ML4CHsurvey}.
Learning-based approaches find a mapping between 2D representations and 3D geometry, which is the inverse of the classic image formation process. The approach is used for reconstructing from information-rich images and low-information representations like sketches.
\subsection{3D Reconstruction from RGB Images}
In the past, numerous learning-based approaches have focused on generating 3D shapes from either single or multiple RGB images. These methods aim to capture various properties, including contours, depths, shadows, and textures.\\
One widely used approach for reconstructing a 3D scene from a single image is neural rendering \cite{park, RPMNet, haozhe}. By establishing correlations between the 2D image and the 3D geometry, this approach employs a deep neural network to generate a novel view of the scene. Training the neural network on a dataset containing pairs of 2D images and 3D models enables it to gain a more comprehensive understanding of the relationships between these two domains, i.e. the image formation process. As a result, the network becomes capable of generating new views of the scene that align with the information provided by the original 2D images. \\
Shape-from-shading uses a neural network to learn the mapping between image features and 3D geometry by analyzing shading and lighting cues in a single image.
One such approach is presented in \cite{jiaming}. Through evaluation on the \href{https://zju3dv.github.io/neuralrecon-w/}{Heritage-Recon} dataset, the authors showcased the superior performance of this approach in terms of both efficiency and accuracy. Notably, it outperformed classical reconstruction methods as well as neural reconstruction methods. 
Another promising idea is the learning of shape priors. By combining the explicit prior knowledge of 3D shapes with the rendering process, this approach enhances the accuracy and reliability of the reconstruction process \cite{latent, reproj}. In \cite{paul} a comprehensive prior model of 3D shapes is learned and integrated with a renderer to model the image formation process. This involves designing a probabilistic generative model that captures the entire image formation process. It includes independent sampling of the 3D mesh's shape, pose, and incident lighting, followed by the generation of a 2D rendering based on these samples.\\
A self-supervised method for 2D-to-3D shape translation using unpaired examples is proposed by \cite{sist}. It utilizes an image generation network and a shape reconstruction network trained on features from both domains. The generative network's output images train the shape reconstruction network, enabling the learning of disentangled representations for domain translation. It is effective in reconstructing single view models from a single RGB image, even with unpaired training sets.
\subsection{3D Reconstruction from Sketches}
Photographs offer a lot of detail such as lighting and texture in addition to general shape. In cultural heritage, depictions are often degraded. Therefor, we look into algorithms working with data providing less information, such as sketches.
\cite{srfs} uses a decoder-encoder architecture to create multi-view images for depth and normals, which are then fused to generate a 3D point cloud. \cite{manmade} forecast parametric 3D shapes using single bitmap sketches of man-made objects. The method combines a convolutional neural network (CNN) with a generative adversarial network (GAN) to facilitate sketch-based 3D modeling. However, this and other works 
focus on objects of rather low complexity. 
\subsection{2D Image Processing}
To convert sketches into photographs, the above mentioned GANs have been widely used for image-to-image translation \cite{style, GAN, wggan}. However, GANs face the challenge of balancing the training of both generator and discriminator, which can lead to instability in the training process. By contrast, another kind of generative models, diffusion models only require one model to train and are easier to optimize, making them a more stable option. As a result, approaches like Pretraining is All You Need for Image-to-Image Translation (PITI) \cite{piti} and Palette \cite{Palette}, which use diffusion models, have been proposed and have shown excellent performance for image-to-image translation. Palette, as a general framework for image translation, can outperform strong GANs and regression baselines for various tasks without requiring task-specific customization or optimization, making it interesting to our domain. PITI explicitly demonstrates reconsctruction of images from sketches.
And further provides a pretext context to guide the output of the diffusion model.
For our image-to-image translation we adopt PITI \cite{piti}, which we utilize to generate depth and normal maps from pre-processed sketches. Since we have depth and normal images for many other sculptures as part of the dataset presented in \cite{stucco}, the training process for this task is straightforward and does not require any real-world images.
\subsection{Previous Work}
We build upon a previous work \cite{stucco}, which already showed the reconstruction using medieval sketches, but drastically increase the expressiveness of the AI models used. We list the differences in detail:
\begin{itemize}
    \item While \cite{stucco} on an encoder-decoder architecture to learn a useful prior of artifact appearances. We use GAN and diffusion models, which outperform the former in terms of visual quality, fidelity, and variety of outputs. Further, diffusion models enable the incorporation of powerful prior knowledge using language models and text prompts.
    \item Our solution is different from \cite{stucco}, as it maps input to multiple outputs, rather than a single output.
    \item \cite{stucco} only estimates a depth image, while our solution gives full 3D.
    \item Our method uses photographs of walls directly as input and automatically pre-processes them, in contrast to \cite{stucco} using intermediate hand-drawn sketches based on wall paintings
\end{itemize}
\subsection{Image inpainting}
Image inpainting is a computer vision technique for filling in missing or damaged parts of an image, used in applications such as image restoration, data recovery, and digital painting. The goal is to create a complete and accurate image that resembles the original scene.
One of the inpainting methods we use in our approach is \cite{medfe}. Here the authors propose a encoder-decoder network for image inpainting by using features extracted from a CNN and introduce a feature equalization method to make structure and texture consistent with each other. 
The other inpainting method used is \cite{misf}, which interprets the problem as a image-level predictive filtering task and explore the challenges and advantages this brings with it. The authors propose a solution built upon two network branches, a kernel prediction branch (KPB) and a semantic image filtering branch (SIFB), both of which are combined to come up with the final result.

\section{Approach}
\label{sec:approach}
\begin{figure}[h]
    \centering
	\includegraphics[width=0.35\textwidth]
 {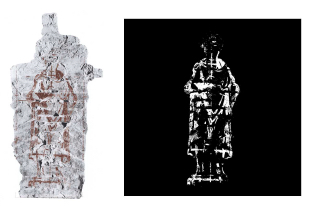}
	\caption
	{In- and output image pairs before and after the automated image pre-processing.}
        \label{fig:duplexEdges}
\end{figure}

Our proposed solution takes an image or sketch and automatically generates an estimated reconstruction of the 3D model. The required pipeline is illustrated in fig. \ref{fig:approach}. \\
We start with automated image pre-processing using OpenCV (b), to enhance the signal to noise ratio and remove noise from the photographs of the sinopia. This step improves the image quality for more accurate analysis and 3D estimation.
The output of the step is shown in fig. \ref{fig:duplexEdges}. 
Howerver the extracted edges from the sketches may have missing parts due to age and mishandling, causing holes in the image. To remove image artifacts and fill in missing parts, an inpainting technique is used.
To solve the inpainting task, a stochastic model is trained on a synthetic dataset of statues presented in \cite{stucco}. The model is trained to reconstruct a good signal after heavy augmentation of the pre-processed images, including the deletion of large parts and sampling of masks with different probability distributions. The end result is a cleaned up, edge-only version of the input.
We use these results as input data, feeding the restored images into a state-of-the-art generative model. As mentioned above, GANs and diffusion models such as StyleGAN 3 \cite{stylegan3}, \href{https://openai.com/product/dall-e-2}{DALL-E 2} and \href{https://stability.ai/blog/stable-diffusion-public-release}{Stable Diffusion}, outperformed encoder-decoder architectures in every relevant metric. Also diffusion models are designed for easy integration of additional inputs, such as text or image masks. For these two reasons we choose to use the diffusion network PITI. We fine-tune our diffusion network to generate normal and depth maps for our sketches (e) yielding images (f). 
In an optional step towards multi-perspective 3D reconstruction (g), additional perspectives of the target geometry could be generated. This should lead to a more complete and accurate reconstruction, creating a more detailed and realistic representation. 
The final step generates point clouds and 3D shapes from depth and normal maps for geometry reconstruction, resulting in a conventional 3D mesh used in 3D applications. Qualitative results are presented in the following section \ref{sec:results}. \\
The integration of additional inputs within our approach can be used for the development of an authoring tool for domain experts to estimate the geometry of a historic object based on a low quality image representation. PITI and its textual understanding can guide the image formation process with additional information provided via text. For example, a damaged historic drawing of a goblet with information that it is made of gold and decorated with gems can be used for reconstruction.
Also, by design, additional inputs may be included, such as segmentation masks, allowing to reconstruct objects from images depicting cluttered scenes, such as the plate in front of Jesus in L’Ultima Cena by Leonardo da Vinci. Our pipeline directly facilities the creation of such a tool, though implementation and testing it is not in the scope of this work. 
\section{Preliminary Results} \label{sec:results}
\begin{figure}
     \centering
     \begin{subfigure}[b]{0.235\textwidth}
         \centering
         \includegraphics[width=\textwidth]{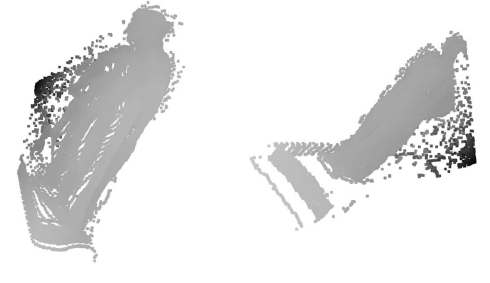}
         \caption{Result of our reconstructed point cloud, estimated from the corresponding depth map.}
         \label{fig:pcd_result}
     \end{subfigure}
     \hfill
      \begin{subfigure}[b]{0.235\textwidth}
         \centering
         \includegraphics[width=\textwidth]{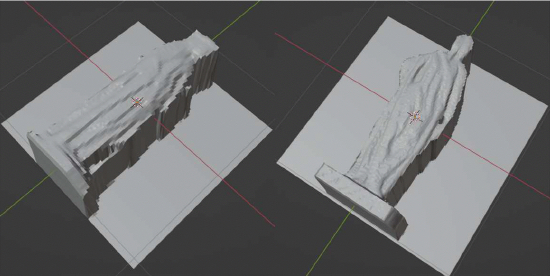}
         \caption{Result of our reconstructed mesh, estimated from the corresponding depth map.}
         \label{fig:mesh_resutl}
     \end{subfigure}
     \caption{Preliminary 3D reconstrution results}
        \label{fig:PreliminaryResults}
    
\end{figure}
Our approach has to generalize to unseen images of historic sinopia drawings. To improve generalization while training on renderings only, we applied pre-processing. We presented the extracted shape of our input data in section \ref{sec:approach} and show qualitative results of a reconstructed point cloud in fig. \ref{fig:pcd_result} and a reconstructed mesh in fig. \ref{fig:mesh_resutl}. These reconstructions are made with their corresponding generated depth map and without generating additional views as outlined as optional above.
As for the resulting quality, both the point cloud and reconstructed mesh are visually appealing. In the sample depicted, we find fine details in the robe and accurate hand positioning. The results transfer to other, similar statues. There is, however, one clear weakness to our current approach: Since we are using the dataset from \cite{stucco} we see a qualitatively good shape reconstructions for statues similar to the training set (humanoid statues from ancient Rome, ancient Greece and the Middle Ages), but less so for unseen categories and objects. 



\section{Conclusion and Future Work}
Our work makes state-of-the-art deep learning algorithms available to experts in the cultural heritage domain, enabling the estimation of 3D shapes of objects with only low quality, low information representations like historic sketches. The diffusion network (PITI) can handle degraded, incomplete, and non-planar materials like stone walls with visible seams.
In conclusion, we demonstrated the reconstruction of 3D geometry in the cultural heritage domain using state-of-the-art generative AI models. In detail we presented three contributions:
First, a review of the relevant literature was conducted to identify approaches for reconstructing 3D geometry from a single input to multiple outputs in the cultural heritage domain and with cultural artifacts.
Second, state-of-the-art generative models, including GANs and diffusion models, have been used to showcase their applicability in the reconstruction of cultural heritage artifacts. These advanced techniques allow for a wider selection of input sources, as demonstrated by the reconstruction process using simple photographs of sketches found on a medieval wall.
Third, an interactive authoring tool, tailored for domain experts in the cultural heritage field, has been proposed. The tool allows experts to reconstruct artifact appearances based on simple inputs like images and text. The tool aims to enhance expert capabilities, enable creative and interpretive processes, and improve the reconstruction of cultural heritage artifacts.\\
Future work includes investigating the use of multiple sources of inputs, such as old photographs, and their impact on overall image quality and fidelity. Next, the effects of prior knowledge introduced via textual prompting shall be explored. Finally, we want to design and implement the proposed tool for and with domain experts, and investigate its usefulness in a study with practitioners in the cultural heritage domain. 

The work presented in this paper has been funded by the European Commission during the project PERCEIVE under grant agreement 101061157.

\bibliographystyle{eg-alpha-doi}  
\bibliography{egbibsample}     

\end{document}